%% file: main.tex
\documentclass[runningheads]{llncs}

\usepackage[T1]{fontenc}
\usepackage{graphicx}
\usepackage{amsmath,amssymb,amsfonts}
\usepackage{booktabs}
\usepackage{multirow}
\usepackage{tabularx}
\usepackage{algorithm}
\usepackage{algpseudocode}
\usepackage[table]{xcolor}
\usepackage{float}
\usepackage{pgfplots}
\usepackage{svg}
\usepackage{wrapfig}
\usepackage{enumitem}
\setlist{nosep}
\usepackage{subcaption}
\usepackage{comment}
\usepackage[hidelinks]{hyperref}
\usepackage[nocompress]{cite}
\usepackage[colorinlistoftodos]{todonotes}

\usepackage{orcidlink}

\makeatletter
\newcommand{\printfnsymbol}[1]{%
  \textsuperscript{\@fnsymbol{#1}}%
}
\makeatother

\newcommand{\vicup}{ViCuP}

\definecolor{savierrow}{RGB}{235,243,248}
\newcommand{\runin}[1]{\par\noindent{\normalfont\bfseries\itshape #1}\enspace} 

\begin{document}
\title{SAVIER: Source-Anchored Visual Inspection and Evidence-Guided Repair for Vietnamese Cultural Text-to-Image Generation}

\titlerunning{SAVIER: Vietnamese Cultural Text-to-Image Generation}

\author{
Gia-Huy Thai\inst{1,2}\orcidlink{0009-0006-6684-5323}\thanks{These authors contributed equally to this work.} \and
Duc-Tien Khong\inst{1,2}\orcidlink{0009-0006-3557-0558}\printfnsymbol{1} \and
Trung-Nghia Le\thanks{Corresponding author.}\inst{1,2}\orcidlink{0000-0002-7363-2610}
}

\authorrunning{Gia-Huy Thai et al.}

\institute{
University of Science, Ho Chi Minh City, Vietnam
\and
Vietnam National University, Ho Chi Minh City, Vietnam
\\
\email{
\{23120008, 23120173\}@student.hcmus.edu.vn,\\  
ltnghia@fit.hcmus.edu.vn
}
}

\maketitle              
\begin{abstract}
We present SAVIER, a training-free, one-pass framework that repairs the depiction of a named Vietnamese cultural entity in a generated image while preserving the scene details requested by the user. The problem is hard because a plausible scene can miss the cues that identify the entity, and prompt-level refinement never inspects the image it produces. SAVIER starts from a Culture-TRIP expanded prompt; a Curator separates source-linked cultural cues from prompt-preservation requirements, a prompt-blind Observer describes the first image, and a Refiner turns the discrepancies into at most three positive instructions before one regeneration with the same seed and a fixed bank of reference photographs. On the 50 bilingual prompts of \vicup{}, evaluated with SDXL and FLUX.1-dev against a frozen, human-verified visual contract, SAVIER raises contract fulfillment from 55.6 to 69.6 and from 59.1 to 67.3 over Culture-TRIP and raises CAIRE cultural relevance on FLUX.1-dev. A reference-only ablation that removes the agent repair shows that SAVIER's photographs account for most of the contract gain; the agent roles add country-level relevance on FLUX.1-dev without lowering adherence to the original prompt under the official VQAScore.
\keywords{Vietnamese cultural concepts \and Cultural text-to-image generation \and Image-aware prompt refinement \and Reference-guided generation}
\end{abstract}
\input{contents/1_intro}
\input{contents/2_related_work}
\input{contents/3_method}
\input{contents/4_dataset_v2}
\input{contents/5_experiments}

\input{contents/discussion}


\begin{credits}
\subsubsection{\ackname} 
This research is funded by University of Science, VNU-HCM, under grant number CNTT 2025-23. Trung-Nghia Le was funded by the Postdoctoral Scholarship Programme of Vingroup Innovation Foundation (VINIF), VinUniversity, code VINIF.2025.STS.14.
\end{credits}
%
%
%
\bibliographystyle{splncs04}
\bibliography{references}
\end{document}

%% file: contents/1_intro.tex
\section{Introduction}

Text-to-image (T2I) models render plausible Vietnamese scenes while missing the cues that identify the cultural object named in the prompt: an image may match the setting yet lack the construction, shape, or arrangement that makes the object recognizable~\cite{liu2024scoft,shi2026culture}. The prompt is also incomplete in a second way. A request for an \emph{ao dai} does not enumerate the garment's distinguishing features, yet it may specify supporting objects, actions, or effects that are not intrinsic to the garment. A repair procedure must therefore satisfy two requirements without conflating them: depict source-supported cues of the focal entity, and preserve the scene details the user asked for.

\vspace{-0.1em}

Existing interventions enter at different points. Culture-TRIP retrieves cultural descriptions to refine the prompt before generation but never inspects the image it produces~\mbox{\cite{jeong2025culture}}\hskip0pt. ImageRAG compares a generated image with its prompt, retrieves references for the detected gaps, and regenerates~\cite{shalev2026imagerag}. FAGER builds verifiable rubrics from proposed facts and reference evidence and uses them for refinement~\cite{lim2026fager}. Image-aware feedback and external visual evidence are therefore not new in themselves. Our question is how to organize a bounded repair of a Vietnamese cultural image around the two requirements above, and how much of the gain is due to the repair rather than to the reference images it uses.

We study this on \vicup{}, a bilingual Vietnamese-English set of 50 prompts, each built around one Vietnamese cultural concept. Our training-free framework, SAVIER, starts from an image generated with a saved Culture-TRIP prompt. A Curator separates prompt-explicit scene requirements from Wikipedia-sourced cultural cues; an Observer describes the initial image without seeing the prompt, the entity, or the sources; a Refiner turns the discrepancies into at most three positive instructions. The generator then runs once more with the same seed, the revised prompt, and a fixed bank of reference photographs. The three roles are functional roles of one multimodal backbone, not separately trained agents.

On both SDXL and FLUX.1-dev, SAVIER raises visual-contract fulfillment over Culture-TRIP and raises country-level cultural relevance on FLUX.1-dev. An ablation that keeps SAVIER's curated photographs but removes the agent repair shows that the photographs are the main source of the contract gain, while the agent roles add country-level relevance on FLUX.1-dev without lowering adherence to the original prompt; on SDXL, SAVIER and the ablation are statistically indistinguishable.

Our contributions are:
\begin{itemize}
  \item SAVIER, a training-free, one-pass repair framework that separates sourced cultural cues from prompt-preservation requirements and fixes evidence and reference images offline.
  \item A Curator-Observer-Refiner design in which each role sees controlled information, with a prompt-blind first observation and bounded positive repair actions.
  \item \vicup{}, 50 bilingual prompts across eight categories, each naming one focal entity, with a source-linked, human-verified visual contract, and an evaluation on two generators that attributes the gain between reference conditioning and agent repair.
\end{itemize}

%% file: contents/2_related_work.tex
\section{Related Work}


\runin{Cultural text-to-image generation.}Multilingual access does not guarantee culturally specific outputs~\mbox{\cite{shi2026culture}}. One line of work changes the model or its data: SCoFT fine-tunes contrastively~\cite{liu2024scoft}, AltDiffusion adapts the text encoder to new languages~\mbox{\cite{ye2024altdiffusion}}, and CultureCLIP trains a discriminative encoder on visually similar but culturally distinct concepts~\mbox{\cite{huang2025cultureclip}}\hskip0pt. Culture-TRIP keeps the generator fixed and enriches the prompt with retrieved cultural descriptions~\mbox{\cite{jeong2025culture}}\hskip0pt. These approaches improve cultural grounding before or during generation, but do not explicitly diagnose cultural requirements from the generated image itself; we use Culture-TRIP's saved expansions as input and begin image-conditioned diagnosis afterward.

\runin{Image-aware refinement and reference conditioning.}A second line inspects a draft image: Idea2Img revises prompts from multimodal feedback~\cite{yang2024idea2img}, SLD corrects objects and layout under LLM control~\mbox{\cite{wu2024self}}, and T2I-Copilot distributes interpretation, generation, and assessment across agents~\cite{chen2025t2i}; MosAIG and ViFA-Council apply agents to multicultural prompts and Vietnamese folk-painting outpainting~\mbox{\cite{bhalerao2026cultures,nguyen2026vifa}}. A third line conditions generation on retrieved images: Re-Imagen and RealRAG retrieve references for rare or fine-grained entities~\mbox{\cite{chen2022re,lyu2025realrag}}, ImageRAG retrieves for gaps found in a first image and regenerates~\mbox{\cite{shalev2026imagerag}}, and FAGER verifies proposed facts against reference images before refinement~\cite{lim2026fager}. SAVIER shares first-image inspection and reference-conditioned regeneration with these methods; its distinct element is a source-anchored text repair that separates user-specified scene details from sourced cultural cues, with an Observer that never sees the prompt and a Refiner that emits at most three positive actions. Our reference-only ablation (Section~\ref{sec:exp_main}) indicates that reference conditioning drives most of the attribute gain and that the added value of the text repair is generator-dependent.

\runin{Cultural datasets and evaluation.}CUBE, CAIRE, CulturalFrames, and CuRe assess country-level awareness, relevance, expectations, and long-tail artefacts~\mbox{\cite{kannen2024beyond,yayavaram2026caire,nayak2025culturalframes,rege2025cure}}, and CULTDIFF evaluates culturally inclusive generation through visual comparisons~\mbox{\cite{bayramli2025diffusion}}\hskip0pt. Vietnamese content appears in MC-18 and When Cultures Meet~\cite{ye2024altdiffusion,bhalerao2026cultures}, and VietFashion targets \emph{ao dai} retrieval~\cite{cao2026vietfashion}. \vicup{} differs in using one annotated focal entity per bilingual prompt with source-linked visual requirements, which enables item-level diagnosis beyond country-level relevance.

%% file: contents/3_method.tex
\section{Proposed Method}


\subsection{Overview and Offline Grounding}
\label{sec:method_sources}

\begin{figure}[t!]
  \centering
  \includegraphics[width=0.8\textwidth]{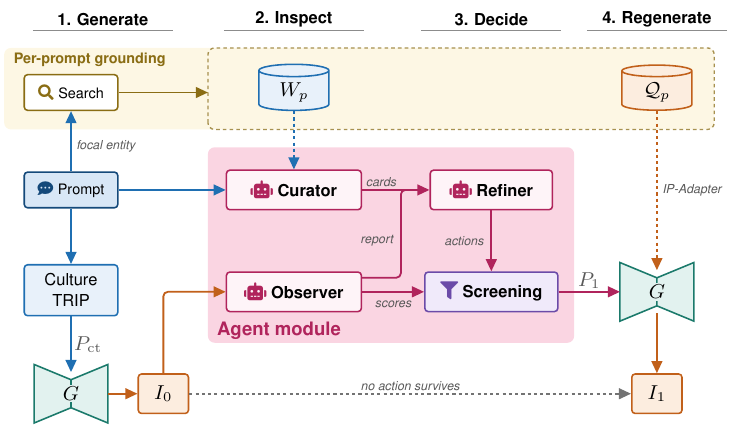}
  \caption{SAVIER in four stages. The Curator reads the prompt and the retrieved passages, the Observer reads only $I_0$, and the Refiner reads their messages but never the pixels. Proposed actions are screened against $I_0$ before the single reference-conditioned regeneration.}
  \label{fig:pipeline}
\end{figure}

Let $p=(p_{\mathrm{vi}},p_{\mathrm{en}})$ be a bilingual request whose focal entity is a Vietnamese cultural concept and $P_{0,p}=p_{\mathrm{en}}$ its English text. We use a saved Culture-TRIP expansion $P_{\mathrm{ct},p}$~\cite{jeong2025culture}, prepending $P_{0,p}$ when it is not already a prefix, as the initial generator input. With generator $G$ and seed $s$,
\begin{equation}
  I_{0,p}=G(P_{\mathrm{ct},p},s).
  \label{eq:initial_arm}
\end{equation}
\vspace{-0.2em}
Before generation, an offline stage links the focal entity to Vietnamese and English Wikipedia pages and stores source-linked text as $W_p$. In parallel, web and Wikimedia Commons search collects candidate photographs; a vision-language filter checks or crops them, and CLIP ranks the retained candidates to form the fixed bank $\mathcal{Q}_p$. From $p$ and $W_p$, an LLM drafts a \emph{visual contract} of observable identity cues, confusable appearances, and source links, which a human verifies and, where needed, edits before generation. The contract is used to index evidence for the Curator and to define the evaluation criteria; the agents never read it, and the photographs enter only the final generator call. SAVIER then diagnoses $I_{0,p}$ and makes exactly one reference-conditioned regeneration (Fig.~\ref{fig:pipeline}).

\subsection{Curator, Observer, and Refiner}

The three roles are role-specific calls to one frozen Mistral Small 3.1 multimodal backbone~\mbox{\cite{mistral2025small31}}, with a fixed message flow and no post-generation evaluator in the loop.

\runin{Curator (C).}From $(p_{\mathrm{vi}},P_{0,p})$ alone, C first writes a \emph{Prompt Preservation Card}: subject, requested attributes, supporting objects, setting, actions, relations, counts, colors, and visible effects. From $W_p$ (at most about 4,000 characters, Vietnamese passages first) it writes a \emph{Cultural Evidence Card} with at most three camera-observable identity cues, two conditional cues, and two positive descriptions that distinguish the target from look-alikes, each with a Vietnamese quotation and a source URL. This is selection within prepared evidence, not new retrieval. A deterministic filter keeps a cue only if its quotation occurs verbatim in the fetched text, its URL belongs to the retrieved pages, and the cue and quotation overlap lexically. This is a provenance check, not a test of semantic entailment. A scene detail such as steam belongs to the preservation card when the prompt requests it, not to the entity's identity.

\runin{Observer (O).}The first O call receives only $I_{0,p}$: no prompt, entity label, contract, or sources. It reports visible subjects, objects, setting, actions, relations, counts, colors, small effects, and uncertain regions without judging cultural authenticity, so that observation is separated from the requirements it is later compared against.

\runin{Refiner (R).}R receives $P_{0,p}$, the Culture-TRIP expansion, both cards, and O's report, but no pixels. It classifies missing prompt details, missing identity cues, contradictions, satisfied details, and uncertain observations, giving prompt-explicit content, especially supporting objects and scene relations, priority over additional cultural cues, and proposes at most three short positive English actions. It may mark misleading phrases of the expansion for removal, never phrases of $P_{0,p}$; actions containing negation or framing changes are rejected.

\runin{Targeted check.}For each proposed action, O receives $I_{0,p}$ and the action as a statement and scores the match on a 0 to 10 scale; actions scoring at least 8 are removed as already present, and free slots are filled with checked identity cues from C's card. This screening step (Fig.~\ref{fig:pipeline}) is a deterministic filter over O's scores, not a fourth agent; it precedes regeneration and does not score $I_1$.

\subsection{Prompt Repair and Regeneration}

Let $\mathcal{A}_p$ be the surviving actions. $P_{1,p}$ keeps $P_{0,p}$ verbatim, removes the approved phrases from the expansion with punctuation cleanup, and appends a preservation sentence naming the card's supporting objects and the actions in $\mathcal{A}_p$. No negative prompt is used. The evaluated output is
\begin{equation}
  I_{1,p}=G(P_{1,p},s;\mathcal{Q}_p), \qquad P_{1,p}=P_{\mathrm{ct},p}\ \text{if }|\mathcal{A}_p|=0,
  \label{eq:final_arm}
\end{equation}
\vspace{-0.2em}
with the same seed and the photo bank applied in every case, including when no action survives. SDXL~\cite{podell2023sdxl} uses up to two photographs through IP-Adapter Plus~\cite{ye2023ipadapter}; FLUX.1-dev~\cite{bfl2024flux1dev} uses one through an XLabs IP-Adapter. $I_1$ is a new generator output, not a pixel edit or a gated selection. Because photographic conditioning changes in every case, the $I_0$ to $I_1$ comparison measures the complete pipeline; the reference-only ablation in Section~\ref{sec:exp_setup} isolates the effect of the revised text.


%% file: contents/4_dataset_v2.tex
\section{ViCuP Dataset}
\label{sec:dataset}

\vicup{} (\textbf{Vi}etnamese \textbf{Cu}ltural \textbf{P}rompts) is a bilingual Vietnamese-English set of 50 prompts, each built around exactly one Vietnamese cultural concept, with 50 distinct concepts. Restricting a prompt to one focal concept keeps an observed failure attributable to that concept. The prompts cover eight categories adapted from prior cultural benchmarks~\mbox{\cite{kannen2024beyond,bayramli2025diffusion,khanuja2024image}}: food and drink (9), clothing (8), arts and music (8), everyday life and trades (6), architecture and landmarks (6), customs and rituals (5), landscape (4), and festivals (4). Every category is small, so we report results over the whole set rather than per category. At 50 prompts, \vicup{} is a focused probe rather than a broad survey. It asks whether a generator can render the visual identity of individual Vietnamese concepts.

Two principles guided construction. \emph{Depictability}: prompts favor concepts with a clear visual form, and for activities such as performances they name visual cues (instruments, attire) through which the activity is recognized. \emph{Naming over describing}: a concept is referred to by its name (``a Cham tower'') rather than by its attributes (``a red-brick Cham tower''), while contextual details such as setting or time of day are kept, so that the evaluation asks whether a model knows what a concept looks like rather than whether it follows a description. For each category, a large language model proposed candidate entities and prompts, which we filtered and revised under these principles. Each prompt carries the human-verified visual contract of Section~\ref{sec:method_sources}: scored requirements with importance weights, conditional requirements, and descriptions of confusable traits. The 50 contracts contain 150 scored requirements (two to four per prompt), six conditional requirements, and 115 descriptions of confusable traits (two or three per prompt), linked to 55 distinct source URLs; English prompts are 7 to 18 words long (mean 12).

%% file: contents/5_experiments.tex
\section{Experiments}

\subsection{Experimental Setup}
\label{sec:exp_setup}

We evaluate the 50 \vicup{} prompts with SDXL~\cite{podell2023sdxl} and FLUX.1-dev~\cite{bfl2024flux1dev} (hereafter FLUX). All arms use seed 5000 and one $1024\times1024$ image per prompt; SDXL runs 30 steps at guidance 5.0 with up to two photographs through IP-Adapter Plus at scale 0.5, FLUX 28 steps at guidance 3.5 with one photograph through the XLabs IP-Adapter at scale 0.6. We compare the original prompt $A_p=G(P_{0,p},s)$, the Culture-TRIP expansion $I_{0,p}$ (Eq.~\ref{eq:initial_arm}), and SAVIER $I_{1,p}$ (Eq.~\ref{eq:final_arm}). To attribute the gain, we add a \emph{reference-only} ablation $I_{\mathrm{ref},p}=G(P_{\mathrm{ct},p},s;\mathcal{Q}_p)$: the Culture-TRIP prompt with the same seed and photographs and no agent call, so that $I_{\mathrm{ref}}$ and $I_1$ differ only in the text given to $G$. When no action survives (8 SDXL and 5 FLUX prompts), $I_1$ coincides with $I_{\mathrm{ref}}$.

\subsection{Evaluation Protocol}
\label{sec:exp_evaluation}

\runin{Cultural relevance.}The original CAIRE implementation~\cite{yayavaram2026caire} gives five-point relevance scores for the labels Vietnam (CAIRE-VN, higher is better) and China (CAIRE-CN, reported to expose cross-cultural attribution; lower is better). CAIRE assesses country-level relevance, not whether the focal object has its required attributes. The original-prompt and Culture-TRIP images were scored in the initial CAIRE run and the SAVIER and reference-only images in a later run of the same code, which scores each label in a separate call, so the two runs are comparable.

\runin{Contract-item judgments.}Qwen2.5-VL-7B-Instruct~\cite{bai2025qwen25vl} judges each image against the individual statements of the frozen contract, without the arm label or transcripts, on hashed filenames in shuffled order; \textsc{yes}, \textsc{partial}, and \textsc{no} map to $1$, $0.5$, and $0$. Because the contract also indexes the Curator's evidence, these scores quantify fulfillment of specified criteria rather than independent human judgments. Let $R_p$ be the unconditional, camera-observable requirements with weights $w_i$ and $y_i(I)$ the judgment for image $I$; let $F_p$ be the confusable descriptions with judgments $z_j(I)$. Then
\begin{equation}
\mathrm{VCFS}_p(I)=100\frac{\sum_{i\in R_p}w_i y_i(I)}{\sum_{i\in R_p}w_i},\qquad
\mathrm{CCR}_p(I)=100\frac{\sum_{j\in F_p}z_j(I)}{|F_p|}.
\label{eq:vcfs_ccr}
\end{equation}
\vspace{-0.3em}
VCFS is the weighted percentage of required evidence present (higher is better); conditional requirements are assessed separately and excluded items do not enter the score. CCR is the average visibility of confusable traits (lower is better; undefined when $F_p$ is empty); it does not classify a whole image as depicting a particular alternative entity.

\runin{Prompt adherence.}We report the official VQAScore~\mbox{\cite{lin2024vqascore} }\hskip0ptwith its \texttt{clip-\allowbreak flant5-\allowbreak xxl} backbone, the probability of \textsc{yes} to "Does this figure show $P_{0,p}$? Please answer yes or no." Neither CAIRE nor VQAScore reads the contract, so both are independent of the evidence given to the Curator; CAIRE shares the Qwen2.5-VL-7B scorer with the contract judge, but neither shares the agents' backbone.

Distribution-level measures such as FID cannot tell whether a particular prompt's entity is depicted correctly, so the four metrics answer complementary questions rather than forming one quality score. All numbers are prompt-level means per generator. Paired differences are computed within prompts, with 95\% percentile intervals (in brackets) from 2,000 prompt-level bootstrap resamples; one seed is evaluated, so the intervals describe variability across prompts, not across seeds.

\subsection{Results}
\label{sec:exp_main}

\begin{table}[t!]
\centering
\caption{Results on \vicup{} (50 prompts; prompt-level means; CAIRE on a five-point scale, VQA the official VQAScore, VCFS and CCR in percent). Only SAVIER rows use reference photographs; the reference-only ablation is in the text.}
\label{tab:main_results}
\small
{\setlength{\tabcolsep}{2.8pt}
\renewcommand{\arraystretch}{1.12}
\begin{tabular}{@{}lccccc@{}}
\toprule
\textbf{Configuration} & \textbf{CAIRE-VN} $\uparrow$ & \textbf{CAIRE-CN} $\downarrow$ & \textbf{VQA} $\uparrow$ & \textbf{VCFS} $\uparrow$ & \textbf{CCR} $\downarrow$ \\
\midrule
SDXL & 2.66 & 2.46 & \textbf{0.861} & 52.9 & 20.8 \\
SDXL + Culture-TRIP & 3.00 & 2.54 & 0.791 & 55.6 & 15.3 \\
\rowcolor{savierrow}
\textbf{SDXL + SAVIER} & \textbf{3.36} & \textbf{2.36} & 0.805 & \textbf{69.6} & \textbf{15.2} \\
\midrule
FLUX & 2.50 & \textbf{2.54} & 0.832 & 44.8 & \textbf{13.3} \\
FLUX + Culture-TRIP & 2.50 & 3.08 & \textbf{0.839} & 59.1 & 15.0 \\
\rowcolor{savierrow}
\textbf{FLUX + SAVIER} & \textbf{3.00} & 2.88 & 0.833 & \textbf{67.3} & 14.2 \\
\bottomrule
\end{tabular}
}\end{table}

The generators respond differently to prompt expansion (Table~\ref{tab:main_results}). On SDXL, Culture-TRIP raises CAIRE-VN but barely VCFS; on FLUX it raises VCFS while CAIRE-VN stays at 2.50. SAVIER raises both on both generators. Target-cue fulfillment and country-level relevance need not move together. Confusion results are less uniform: SAVIER lowers CAIRE-CN relative to Culture-TRIP on both generators, but CCR changes only slightly, and the original FLUX prompt retains the lowest CAIRE-CN and CCR within its group. Paired differences $I_1-I_0$ are $+14.0$ VCFS ($[+6.0,+21.9]$) and $+0.36$ CAIRE-VN ($[-0.16,+0.88]$) on SDXL, and $+8.3$ VCFS ($[+1.0,+15.5]$) and $+0.50$ CAIRE-VN ($[+0.10,+0.94]$) on FLUX; CCR intervals include zero.

\vspace{-0.1em}

\runin{Attribution of the gain.}Without the agent roles, the reference-only ablation reaches 72.3 VCFS on SDXL and 65.3 on FLUX, so SAVIER's photographs account for most or all of the contract gain over $I_0$ on both generators. Beyond this ablation, the agent roles raise CAIRE-VN on FLUX by $0.28$ ($[+0.04,+0.54]$), the only one of ten SAVIER minus reference-only intervals that excludes zero, without correction for multiple comparisons; the VCFS ($+2.1$, $[-4.0,+8.2]$) and CCR ($-2.7$, $[-8.3,+2.7]$) differences are within noise. On SDXL the two arms are statistically indistinguishable on every metric (VCFS $-2.7$, $[-7.4,+1.3]$; CAIRE-VN $-0.28$, $[-0.58,+0.02]$). The lower CAIRE-CN of SAVIER relative to Culture-TRIP is shared by the ablation and is therefore attributable to the photographs. Official VQAScore varies by at most 0.03 across Culture-TRIP, the ablation, and SAVIER on both generators, so the repair does not lower adherence to the original prompt. For 4 of the 50 prompts per generator the Curator returned no parsable evidence card and the Refiner acted on the preservation card alone; excluding them changes no conclusion.

\runin{Runtime.}On one A100 80\,GB with both models resident, caches disabled, and model loading and offline grounding excluded (five prompts, three repeats after a warm-up), the Curator takes $53.4\pm28.1$\,s per prompt, the Observer $12.1\pm0.8$\,s, and the Refiner $13.8\pm1.6$\,s on SDXL, with nearly identical values on FLUX; the reference-conditioned generation takes $14.5\pm1.6$\,s on SDXL and $41.3\pm2.4$\,s on FLUX, against $4.5$ and $34.2$\,s for a plain generation. SAVIER thus adds $93.8\pm26.8$\,s (SDXL) or $119.0\pm27.9$\,s (FLUX) to an existing $I_0$. The Curator depends only on the prompt and the cached passages, so its cards are reused across generators and seeds; the per-image agent cost is the Observer and Refiner, about 26\,s. Capping the evidence card at 1,200 tokens produced identical cards on all benchmark units and reduced Curator time to $45.8\pm14.4$\,s, so most of its variance comes from overlong outputs.

\begin{figure}[t]
  \centering
  \includegraphics[width=0.9\textwidth]{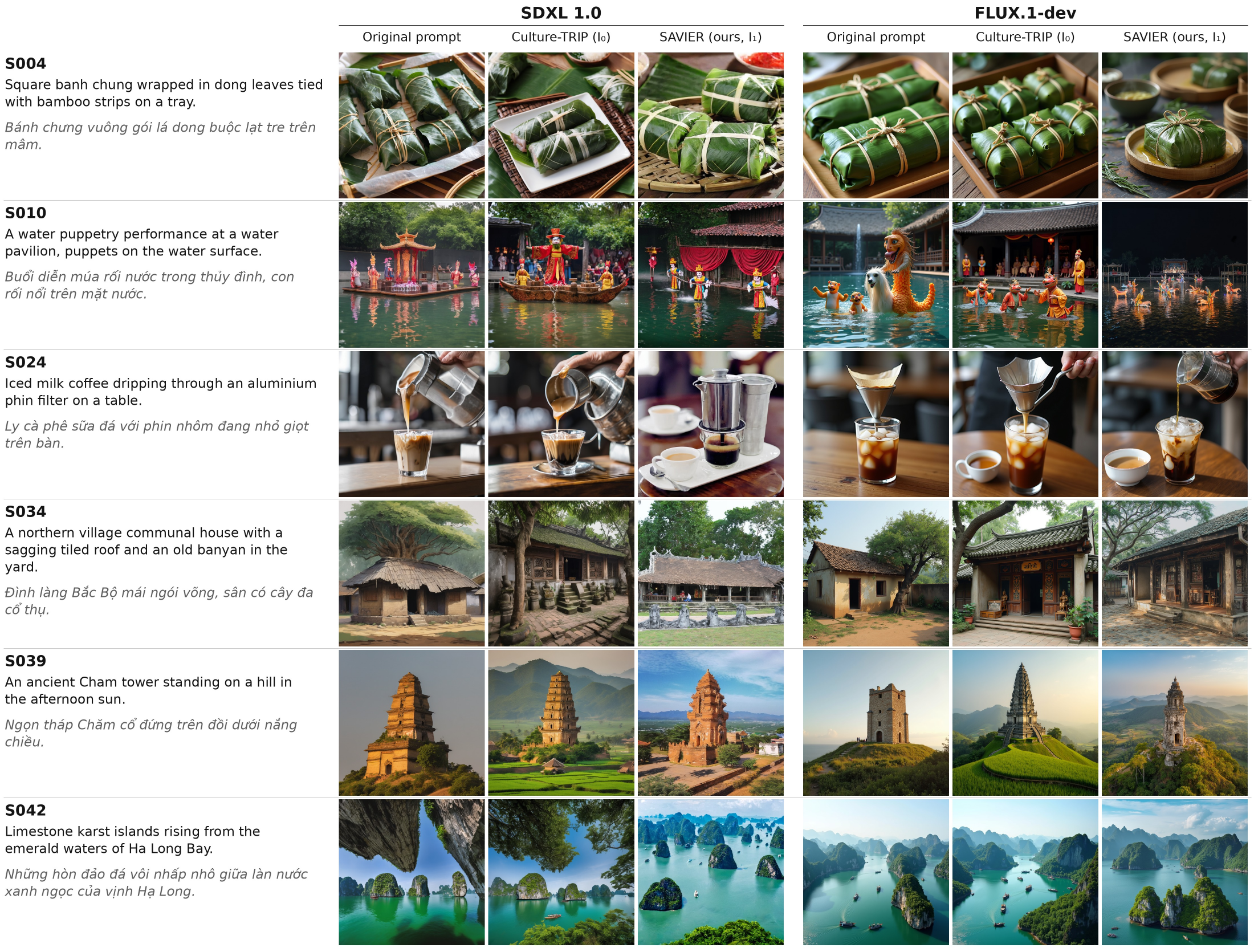}
  \caption{Six prompts with the original prompt, Culture-TRIP ($I_0$), and SAVIER ($I_1$) per generator; bilingual prompts at left. The set includes improvements and regressions and is illustrative, not a random sample.}
  \label{fig:qualitative}
\end{figure}

\runin{Qualitative examples.}Figure~\ref{fig:qualitative} shows cases that the mean scores conceal. On SDXL, S010 gains a recognizable water-puppet scene (VCFS 50 to 100) and S039 a red-brick Cham tower (66.7 to 100), while S034 loses the communal-house roof (100 to 75). On FLUX, S024 falls from 75 to 25 as the phin drip is not preserved and S039 from 66.7 to 33.3 as the tower becomes less diagnostic. A diagnosed cue is not always rendered, and regeneration can lose correct details.


%% file: contents/discussion.tex



\section{Discussion and Conclusion}

SAVIER alters only the conditioning of a frozen generator, so its repairs are bounded by what the generator can render, and one regeneration may be insufficient when the initial depiction is far off; relative to $I_0$, VCFS falls on 8 SDXL and 10 FLUX prompts for SAVIER against 7 and 8 for the reference-only ablation, so most regressions stem from reference-conditioned regeneration rather than from the agents' text. The gains are measured on one seed and 50 prompts, and the agent contribution beyond reference conditioning is small on FLUX and indistinguishable from zero on SDXL. The visual contract is verified by a human before generation, and the contract's source links decide which passages the Curator reads first, so new entities require this offline preparation. The targeted check shares the Refiner's backbone and is a self-consistency filter, not independent verification; human judgments of the repaired images remain future work. The design assumes a single focal entity; multi-entity prompts are unevaluated.

SAVIER separates source-supported cultural cues from user-stated scene details and repairs an observed first image with one bounded, reference-conditioned regeneration. On the 50 \vicup{} prompts it improves contract fulfillment over Culture-TRIP on both SDXL and FLUX without lowering adherence to the original prompt. The reference-only ablation locates most of that gain in SAVIER's photographs and the remaining, generator-dependent benefit of the agent roles in country-level relevance on FLUX. Determining when text-side repair adds to reference conditioning, and extending the evaluation to multiple seeds, human judgments, and multi-entity prompts, are the next steps.